\documentclass[runningheads]{llncs}
\usepackage{graphicx}
\usepackage{hyperref}       % hyperlinks
\usepackage[table,svgnames]{xcolor}

\usepackage{url}            % simple URL typesetting
\usepackage{booktabs}       % professional-quality tables
\usepackage{multirow}
\usepackage{nicefrac}       % compact symbols for 1/2, etc.
\usepackage{microtype}      % microtypography
\usepackage{bbding}
\usepackage{makecell}

\usepackage{amsmath}
\usepackage{amssymb}

\newcommand{\vect}[1]{\mathbf{#1}} % bold, not italic, vectors
\newcommand{\T}{\ensuremath{^{\text{T}}}}

\begin{document}
\title{Mixing Histopathology Prototypes into Robust Slide-Level Representations for Cancer Subtyping}
\titlerunning{Mixing histopathology prototypes for cancer subtyping}

\author{%
Joshua Butke\inst{1} \and % J.B. is corresponding author!
Noriaki Hashimoto\inst{2} \and
Ichiro Takeuchi\inst{2,3} \and
Hiroaki Miyoshi\inst{4} \and
Koichi Ohshima\inst{4} \and
Jun Sakuma\inst{2,5}
}%

\institute{%
Machine Learning and Data Mining Lab, University of Tsukuba, Japan \and
RIKEN Center for Advanced Intelligence Project, Japan \and
Department of Mechanical Systems Engineering, Nagoya University, Japan \and
Department of Pathology, Kurume University, Japan \and
Department of Computer Science, Tokyo Institute of Technology, Japan\\
}

\authorrunning{J. Butke et al.}

\maketitle  % typeset the header of the contribution
\begin{abstract}
Whole-slide image analysis via the means of computational pathology often relies on processing tessellated gigapixel images with only slide-level labels available.
Applying multiple instance learning-based methods or transformer models is computationally expensive as, for each image, all instances have to be processed simultaneously.
The MLP-Mixer is an under-explored alternative model to common vision transformers, especially for large-scale datasets. Due to the lack of a self-attention mechanism, they have linear computational complexity to the number of input patches but achieve comparable performance on natural image datasets.
We propose a combination of feature embedding and clustering to preprocess the full whole-slide image into a reduced prototype representation which can then serve as input to a suitable MLP-Mixer architecture.
Our experiments on two public benchmarks and one inhouse malignant lymphoma dataset show comparable performance to current state-of-the-art methods, while achieving lower training costs in terms of computational time and memory load. Code is publicly available at \url{https://github.com/butkej/ProtoMixer}.

\keywords{Clustering  \and MLP-Mixer \and Computational Pathology}
\end{abstract}
\section{Introduction}
Whole-slide images (WSIs) are digital files of pathology tissue glass slides, often with gigapixel resolution. The advent of digital slide scanning technology and their increasing ubiquity in many medical facilities, lead to the amount of available WSIs rising drastically.
On the basis of data availability there has been a fast advancement in the field of computational pathology (CPATH), applying various computer vision methods for patient specimen (e.g. tissue or individual cells) analysis \cite{cui2021artificial}.
Currently, machine learning methods in the CPATH community are often based on multiple instance learning (MIL), in which each individual WSI is regarded as a bag of thousands of patched instances with a single, patient-level, coarse-grained label \cite{dietterich1997solving,ilse2018attention}. 
%In practice, each bag is then loaded into memory, and due to the varying amount of instances per bag, feed to a deep MIL model separately. 
To improve state-of-the-art performance for subtyping classification problems there are various additions to the base MIL framework of embedding all instances per bag into a feature representation and then classifying a pooled global feature vector, e.g. hard negative mining \cite{butke2021end} or instance-level clustering to constrain and refine the feature space during training \cite{lu2021data}.

\begin{figure}[ht]
    \centering
    \includegraphics[width=\linewidth]{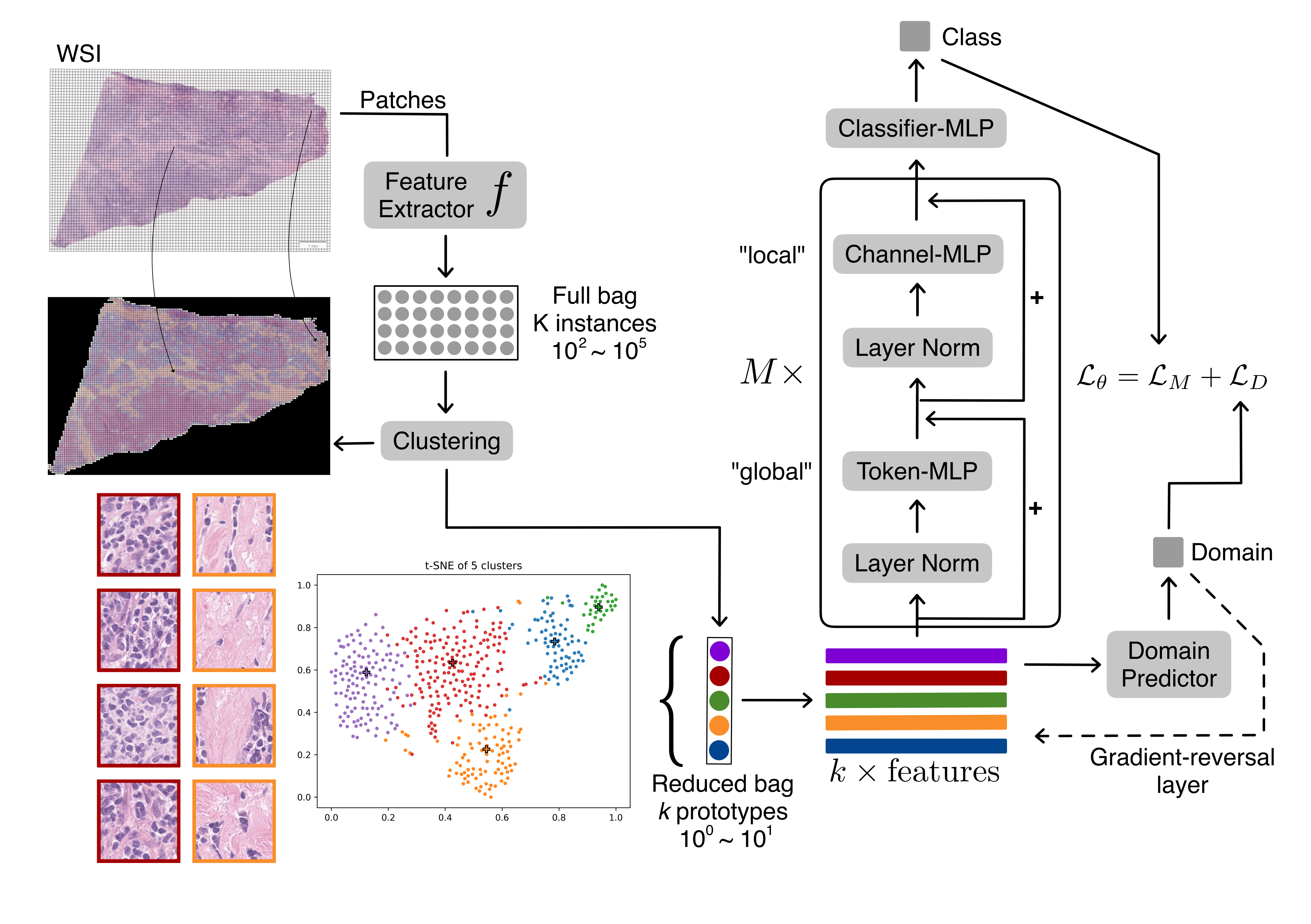}
    \caption{Overview of our proposed end-to-end framework combining feature embedding and clustering of WSIs into reduced prototypes representations (\textbf{left} side) and then feeding the resulting input table into a domain-adversarial Mixer model (\textbf{right} side).}
    \label{fig:pathomix}
\end{figure}

Transformer-based models, such as Vision Transformer \cite{dosovitskiy2020image}, have quickly gained popularity in most of the computer vision community but their application to CPATH image analysis is still lacking as they are constrained by their fixed input sequence length as well as their prohibitive computational cost due to the self-attention mechanism.
The MLP-Mixer model \cite{tolstikhin2021mlp} was proposed as a step away from the inductive biases of convolution in CNNs as well as self-attention found in transformer-based models. 
Instead of those, the model uses only multilayer perceptrons (MLP), the core building block of most neural networks, and applies them either independently to image patches or across patches, effectively operating on a table of "patches $\times$ channels" and its transposition. 
Despite this simplicity, it showed state-of-the-art performance for natural image classification while enabling faster inference and lower computational complexity.
However, the Mixer architecture is originally designed for $16\times16$ patches from a $224\times224$ image, similar to the Vision Transformer. Applying Mixer within the MIL framework on bags with tens of thousands of patches is not computationally possible.
Thus, we pose the question, if it is possible to apply the Mixer architecture to the CPATH realm of gigapixel images, with the goal of competitive predictive performance while achieving faster training times and lower memory overhead, by reducing a bag down to its essential, describing elements. % and then combine this reduced dataset with the model. 
%Therefore we hypothesize:
%    \begin{itemize}
%        \item embedding patches with a standard pretrained convolutional neural network gives good representation of a 2D patch into 1D feature space
%        \item using a clustering algorithm on these embeddings allows us to group patches by similarity
%        \item reducing the number of instances per bag by only keeping the centroid (or prototype) of each cluster decreasing the amount of data to handle by several magnitudes ($10^5$ to $10^1$ instances per WSI)
%        \item the MLP-Mixer architecture is suited to operate on these reduced bags and can achieve good classification performance for cancer subtyping problems.
%    \end{itemize}

Therefore we hypothesize that it is not necessary to use all patches of a given bag, but it is sufficient to use information from the most representative patches.
We propose a solution, called ProtoMixer, by first embedding each patch into a high-dimensional feature space and then use clustering to group patch embeddings by similarity. By only keeping the centroid (or prototype) of each cluster we can reduce a dataset by several orders of magnitude per WSI ($10^5 \mapsto 10^1$ instances).
Compared to current approaches based on MIL this would offer reduced training times due to vastly lower memory overhead (only loading precomputed prototypes from disk) and faster inference speed, even with a larger parameter budget as the input data space is much smaller than typical bags in the MIL setting, while still achieving good cancer subtyping accuracy.%state-of-the-art cancer subtyping accuracy.

%To summarize, the main contributions of our approach are:
%    \begin{itemize}
%        \item  applying a MLP-Mixer-like architecture to the domain of computational pathology for the first time
%        \item clustering feature embeddings of tissue patches into robust prototypes; and
%        \item testing a novel end-to-end framework for carcinoma subtyping in this setting.
%    \end{itemize}
    
% why hard to train and expensive?

\subsection{Related works}
\subsubsection{Attention-based MIL}
Multiple instance learning is the major framework for gigapixel pathology image analysis, especially since the introduction of an attention-like mechanism (ABMIL) in the form of weighted averaging of instances per bag, which takes over the necessary pooling operation \cite{ilse2018attention}.
While not equivalent to full self-attention as found in transformer models (which still is to costly due to the quadratic computational complexity on bags with large amounts of instances), weighted averaging offers individual scores for each instance in a bag, enabling interpretability as well as many further training modifications \cite{li2023deep}.\

One of the most prominent, state-of-the-art applications currently is CLAM (Clustering constrained Attention MIL) \cite{lu2021data}. CLAM introduced preextracting feature vectors for each patch in a WSI with a pretrained CNN, as well as a multi-class subtyping capable ABMIL. Further enhanced by instance-level clustering within the model to encourage the learning of class specific features, a SVM loss is used together with pseudo-generated instance labels from attention scores.

\subsubsection{ReMix Framework}
Similar to our proposed method, the ReMix framework \cite{yang2022remix} first \textbf{re}duces a dataset by embedding patches with a pretrained feature extractor and clustering them and then \textbf{mix}es new, augmented bags from the resulting prototypes.
Mixing in this case refers to four different augmentation methods.
They found that clustering works as, when learned properly, the representation space is shown to be meaningful and distance metrics can show similarity between related patches.
From this setting, the authors subsequently apply various MIL methods, as the framework is model agnostic in this regard and could demonstrate superior performance when compared to their chosen baselines as well as vastly reduced training costs due the generally low parameter count of MIL method and the reduced dataset size.

%%%%%%%%%%%%%%%%%%%%%%%%%%

\section{Method}
This work follows standard multiple instance learning conventions, in which a dataset is composed of $L$ bags and formulated as $D = [(B_i, Y_i)]_{i=1}^L$, where $B_i = \{X_K\}_{j=1}^{L_K}$ is the $i$-th bag consisting of $K$ instances (e.g. tessellated patches extracted from a WSI) and a singular bag label $Y_i$.
%\subsection{Clustering}
\subsubsection{Prototype representation}
We reduce a dataset by combining feature embedding with clustering. For each bag $B_i$ in our dataset and all $K$ instances within we compute a high-dimensional feature vector $\vect{x} \in \mathbb{R}^{1\times N}$ with a pretrained model, where $N$ is the embedding dimension.
Afterwards, any suitable clustering method can be applied to reduce the dataset size. Here we stick with $k$-means clustering \cite{lloyd1982least}. This enables the construction of a reduced bag, consisting only of the cluster prototypes (or centroids), overall compressing the dataset by several orders of magnitude ($10^5 \mapsto 10^1$).

\subsubsection{Intuitions}
We argue that the proposed data preprocessing still offers enough information at the prototype level for the Mixer model to build robust whole-slide-level representations for cancer subtyping tasks. Here, we offer several intuitions on its effectiveness as observed in our experiments:
\begin{itemize}
    \item \textbf{patch-level} By embedding patches to feature vectors we are able to extract information at the lowest level that is often repeated (similar patches).
    \item  \textbf{local-level} Clustering all patches and only keeping the prototypes represents local structures such as tissue component classes (e.g. connective tissue). By mixing at the channel-level, the model can learn about each prototype component.
    \item \textbf{global-level} Finally, there is the token-mixing that acts similar to established self-attention operation and disperses prototype information across different channels to learn a holistic view at the whole-slide level.
\end{itemize}
WSIs often suffer from tissue imbalance, wherein a large number of similar patches convey redundant information, potentially overshadowing minority patches. Breaking the bag down into prototypes bridging the numerical gap between majority and minority patches, we mitigate this problem to some extent. Additionally, employing the mean of embeddings of similar patch groups enhances tissue representation by providing a less noisy depiction.
%We argue that this combination and interplay of Mixer layers is similar to actual pathologist practice as also investigated in other CPATH works \cite{hashimoto2020multi,yang2022remix}. Clinical professionals quickly assess the tissue slide at multiple magnification levels and zoom around the slide visually.

\subsubsection{MLP-Mixer}
Following, we briefly describe the original backbone of MLP-Mixer and how it was adapted to reduced bags of prototypes.
MLP-Mixer stacks $M$ isotropic Mixer layers of the same size and with the exact same internal configuration. Each consists of a token-mixing MLP and a channel-mixing MLP.
Other classic components include: skip-connections, dropout, and a fully-connected classifier head (cf. Fig. \ref{fig:pathomix}).
Originally developed for computer vision tasks, input images are tessellated into non-overlapping patches and then unrolled into feature vectors by a per-patch fully connected layer \cite{tolstikhin2021mlp}.

Since we already obtained feature vectors per patch via the previous embedding step and then reduce our bag of instances to $k$ prototypes we automatically gain the desired input table of $\vect{X} = [x_1,...,x_k] \in \mathbb{R}^{k\times N}$, where $k$ is the chosen amount of clusters to compute and $N$ is the feature dimensionality.
The first MLP block is token-mixing and operates on columns of $\vect{X}$ (so on the transposed input table $\vect{X}\T$); the second is channel-mixing and operates of rows of $\vect{X}$.
$D_S$ and $D_C$ are tunable hidden widths in the token-mixing and channel-mixing MLPs. $D_S$ is selected independently of the number of input patches. Therefore, the computational complexity of the model is linear to the number of input patches, unlike ViT whose complexity is quadratic.
\begin{equation}
    \begin{split}
    &\vect{Y}_{*,i} = \vect{X}_{*,i} + \sigma (\vect{W}_1 \text{LayerNorm}(\vect{X})_{*,i}) \vect{W}_2 \; \text{for}\; i=1...C,\\
    &\vect{Z}_{j,*} = \vect{Y}_{j,*} + \sigma (\vect{W}_3 \text{LayerNorm}(\vect{Y})_{j,*}) \vect{W}_4 \; \text{for}\; j=1...S
    \end{split}
\end{equation} 
Thus, each MLP block contains two fully-connected layers ($\vect{W}$) with a GELU nonlinearity between them ($\sigma$) \cite{hendrycks2016gaussian}.
After all $M$ Mixer layers, $k$ features are generated and average-pooled into a global vector representation of size $N$. A final Classifier-MLP computes the class logits. 

%\subsection{Mixing on the slide-level}
%\subsection{Mixing on the patch-level}
\subsubsection{Domain-adversarial training}
In pathological image datasets, there is often a large variety of staining conditions observed. When similar staining conditions exist only in a specific class or for a small subset of samples, the trained classification model tends to predict class based on the difference in staining conditions.
To decouple the variance in staining intensity from the raw data, stain normalization techniques are often used \cite{macenko2009method}. However, recently domain-adversarial learning has been used to remove the domain information (which also includes the stain variance) from the model representation \cite{ganin2016domain,lafarge2017domain}.
Modern works have shown that domain-adversarial training can augment CPATH algorithms further than standard colour augmentation or stain normalization \cite{hashimoto2020multi}.
Methodically, a second branch that predicts the probability of belonging to a certain domain can be added relatively model-agnostic (cf. Fig. \ref{fig:pathomix}).
\begin{align}
   \text{Optimization of subtyping classifier }\theta_M \leftarrow \theta_M - \lambda_M \frac{\partial \mathcal{L}_M}{\partial \theta_M} \\
   \text{Optimization of domain classifier }\theta_D \leftarrow \theta_D - \lambda_D \frac{\partial \mathcal{L}_D}{\partial \theta_D} \\
   \text{Adversarial update of subtyping classifier }\theta_M \leftarrow \theta_M + \alpha \lambda_D \frac{\partial \mathcal{L}_D}{\partial \theta_M}
\end{align}

The parameters $\theta_M$ are updated for the subtyping task (by minimizing $\mathcal{L}_M$), and with the adversarial update, the same parameters are updated to prevent the domain of origin to be recovered from the learned representation.
We add domain-adversarial training to our framework, expecting that removing the domain information that is still present at the level of patches, embeddings and prototypes, can help to learn robust slide-level representations from just prototype information.
%Therefore, each WSI in the dataset is its own domain that a secondary branch in the model tries to predict. With an adversarial training step the model is updated a second time together with each class prediction to prevent the domain of origin to be recovered from the learned representations.

\subsubsection{ProtoMixer framework}
%Overall, our work proposes a new framework we term ProtoMixer, an application of the main MLP-Mixer architecture to the realm of CPATH with the help of reducing a WSI to highly descriptive prototypes. An descriptive overview can be seen in Fig. \ref{fig:pathomix}: Starting from a WSI we first preprocess all segmented patches by feature embedding with a output feature dimensionality of $N$ and then group these into $k$ defined clusters with the $k$-means clustering algorithm. We only keep the prototypes as the reduced dataset and save them to disk. Each WSI then serves as a $k\times N$ input table into the Mixer model which predicts class logits. A secondary MLP branch is used for domain-adversarial training and computes domain logits in parallel. A gradient-reversal update is then used to decouple the domain information from the learned WSI representation.
Overall, our work proposes a framework we term ProtoMixer (Fig. \ref{fig:pathomix}): From a WSI we extract feature embeddings for all segmented patches with output feature dimensionality $N$ and then group these into $k$ clusters, keeping only the prototypes and saving them to disk. Each WSI serves as $k\times N$ input into a Mixer model.
Domain-adversarial training with a secondary MLP branch decouples domain information from the learned WSI representation.
%%%%%%%%%%%%%%%%%%%

\section{Experiments}
\subsubsection{Implementation details}
%\subsubsection{WSI preprocessing}
Each digital slide is fed to a pipeline for automated segmentation and patch tessellation. A binary mask for tissue regions is generated by thresholding the the image after median blurring for edge smoothing, followed by morphological closing to fill artifacts and holes.
After segmentation, $224\times224$ patches are extracted from the foreground contours at a specified magnification level (e.g. $40\times$ or $20\times$)and saved together with their coordinates. 

%\subsubsection{Patch embedding}
Similar to many established methods in modern computational pathology algorithms, each patch extracted from a WSI is embedded to a 1024-dimensional feature vector by an ImageNet pretrained, truncated ResNet-50 \cite{deng2009imagenet,he2016deep}.
Patch embedding in this way has become standard practice as it allows to process a WSI much easier on commonly available GPUs, leading to generally faster training times and lower computational costs.

The choice of $k$ and thus the number of prototypes per WSI is a very important hyperparameter in this framework. Similar to the ReMix idea, we sweep over a selection of $\{1,2,4,5,6,8,10,12,16\}$ for $k$, for each dataset and analyze meaningful representation of prototypes and assigned clusters in a two-dimensional t-SNE embedding \cite{van2008visualizing} as well as predictive performance. We choose $k=5$ for TCGA-RCC and inhouse-Lymphoma datasets and $k=8$ for CAMELYON.

\begin{table}
\caption{Experimental results of model subtyping performance in terms of macro-F1 score and AUROC on three different datasets. 5-fold cross validation averaged over 5 independent runs. Gray result subset shows available literature results from \cite{xiong2023diagnose} for CAMELYON16 and TCGA-RCC.}\label{tab:results}
\resizebox{\linewidth}{!}{%
\begin{tabular}{lcccccc}
\toprule
\multicolumn{4}{r}{\textbf{Dataset}} \\
& \multicolumn{2}{c}{CAMELYON} & \multicolumn{2}{c}{TCGA-RCC} & \multicolumn{2}{c}{inhouse-Lymphoma} \\
\cmidrule(lr){2-3}                  
\cmidrule(lr){4-5}
\cmidrule(lr){6-7}
& macro-F1 & AUROC & macro-F1 & AUROC & macro-F1 & AUROC    \\
\midrule
\rowcolor{WhiteSmoke}ABMIL \cite{ilse2018attention} & 85.00\tiny±1.80\small & 90.03\tiny±1.30\small & 82.80\tiny±1.80\small & 95.90\tiny±1.00\small & N/A & N/A \\

\rowcolor{WhiteSmoke}TransMIL \cite{shao2021transmil}& 78.70\tiny±3.30\small & 83.80\tiny±4.70\small & 85.00\tiny±0.70\small & 97.20\tiny±0.30\small & N/A & N/A \\

\rowcolor{WhiteSmoke}DTFD-MIL \cite{zhang2022dtfd} &
85.80\tiny±1.70\small & 93.30\tiny±0.90\small & 88.40\tiny±2.80\small & 97.60\tiny±0.90\small & N/A & N/A \\
\midrule
CLAM & 78.72\tiny±0.50\small & 83.33\tiny±0.69\small & 84.36\tiny±0.52\small & 95.90\tiny±0.71\small & 70.18\tiny±0.38\small & 86.41\tiny±0.13\small\\ % done

CLAM+prttyps & 54.47\tiny±0.77\small & 72.30\tiny±0.16\small & 23.76\tiny±0.00\small & 49.22\tiny±0.13\small & 72.29\tiny±0.15\small & 86.07\tiny±0.02\small\\ % done

ReMix+ABMIL & 72.26\tiny±0.91\small & 78.12\tiny±1.04\small & 86.97\tiny±0.22\small & 96.34\tiny±0.38\small & 78.87\tiny±0.25\small & 89.94\tiny±0.39\small\\ % done

\textbf{ProtoMixer} & 72.71\tiny±1.01\small & 76.81\tiny±1.21\small & 89.72\tiny±0.47\small & 97.51\tiny±0.12\small & 82.67\tiny±0.33\small & 93.68\tiny±0.18\small\\ % done
\bottomrule
\end{tabular}
} % close resizebox
\end{table}

%\subsubsection{Mixer model}
For the Mixer model, we choose the dimensionality of fully-connected layers to be $D_S = 1024 \; (= N),\; D_C = 2048$, with the number of stacked Mixer blocks $M = 12$.
The domain-adversarial branch running in parallel during training is a three-layer MLP acting as the domain predictor.
It transforms the input data space into a domain logit, where the size is equal to the number of individual WSIs in a given dataset. A gradient-reversal layer then modifies the total model loss for each epoch. The domain regularization parameter $\lambda$ is updated each epoch as $\lambda = \frac{2}{1+\exp{(-10r)}}$, with $r=\frac{\text{current epoch}\; e}{\text{total epochs}\; E}\;\alpha$, inspired by \cite{hashimoto2020multi}.

\subsubsection{Datasets}
We evaluate our approach on three different datasets. Two are well established benchmark datasets commonly found in current computational pathology research: \textbf{CAMELYON} and \textbf{TCGA-RCC}. 
The TCGA-RCC dataset consists of 884 diagnostic WSIs available from different \emph{The Cancer Genome Atlas} programs for renal cell carcinoma. There are three different subtypes made up of 111 slides from 99 cases for chromophobe carcinoma (TCGA-KICH), 489 slides from 483 cases for clear cell carcinoma (TCGA-KIRC) and 284 slides from 264 cases for papillary cell carcinoma.
The mean number of extracted patches at $40\times$ magnification is 51496.
CAMELYON16 and CAMELYON17 are two competition datasets for breast cancer lymph node metastasis detection \cite{litjens20181399}. As is convention, these two are often combined into a single dataset for binary prediction, consisting of 899 slides (591 negative and 308 positive) from 370 cases.
The mean number of extracted patches per WSI at $40\times$ magnification is 45610.

A third dataset is the \textbf{inhouse-Lymphoma} dataset consisting of 1441 WSIs of malignant lymphoma origin with 3 classes (diffuse large B-cell lymphoma (DLBCL); follicular lymphoma (FL); and Reactive subtype). The mean number of extracted patches per WSI at $40\times$ magnification is 14523.

\subsubsection{Baselines}
We compare against two benchmark baselines in the computational pathology community: CLAM \cite{lu2021data} and the ReMix framework \cite{yang2022remix} with ABMIL as the downstream model. We use the reported settings for these methods.
Since the CAMELYON and TCGA-RCC datasets are popular benchmarks in the CPATH community we show additional literature values for different methods to compare against (cf. Tab. \ref{tab:results}, gray subset).

\subsubsection{Main results}
Experimental results regarding subtyping performance across the presented datasets are shown in Tab. \ref{tab:results}. We perform 5-fold stratified cross validation for all experiments and report the average and its standard deviation of macro-F1 score and area under the receiver operating curve (AUROC) across 5 runs. All models are trained for 150 epochs.

From these main results we can see that our proposed ProtoMixer can attain competitive predictive performance, especially on the TCGA-RCC and lymphoma datasets. 
For the combined CAMELYON dataset there is a drop in performance, which we argue might be related to the low amount of cancer tissue found per WSI. 
%Reducing the data to prototypes might lose too much relevant information here to reliably differentiate positive and negative cases.
Future research could be directed towards performance for WSIs with low amounts of tumor, e.g. sample multiple adjacent prototypes per cluster instead of only one.
An ablation study combining CLAM with the prototypes (CLAM+prttyps) as input fails to achieve good predictive performance, as the CLAM model is not designed to work with only the limited information of prototypes. 
For CLAM and other ABMIL-related methods larger bags are necessary to make the attention mechanism worthwhile. 
The Mixer architecture can work on only a select few prototypes and due to their token and channel mixing operation achieve a good whole slide representation in the absence of attention.

\subsubsection{Training costs}
Standard MIL methods such as CLAM or ABMIL have a relatively large computational overhead as each bag is processed separately and as a whole. CLAM alleviates this already slightly by precomputing feature vectors ahead of model training. In our proposed case, we achieve an even greater reduction outside of training due to only loading $k$ prototypes per WSI. 
Regarding the training costs, we report average duration per training epoch and computational memory load in Tab \ref{tab:costs}.
To ensure fair comparison we perform all budgeting experiments on a single Nvidia A100 GPU with a batch size of 1 and the same number of prototypes ($k=5)$.
Similar to the original ReMix framework \cite{yang2022remix} we also achieve lowered training costs in terms of runtime per epoch during training as well as memory load by utilizing only prototypes as input to our model, even if the parameter budget of Mixer is substantially larger than CLAM or standard attention-based MIL.
\begin{table}[t]
\centering
\setlength{\tabcolsep}{6pt}
\caption{Training costs and run times for different models in comparison to our proposed method on the lymphoma dataset.}\label{tab:costs}
\begin{tabular}{lccc}
\toprule
& Parameters (M) & Memory Load & Seconds/Epoch \\
\midrule
CLAM & 0.9 & 2737.38 MiB & 36.47\\
CLAM+prttyps & 0.9 & 8.43 MiB & 4.80\\
ReMix+ABMIL & 0.5 & 7.46 MiB & 4.49 \\
\textbf{ProtoMixer} & 14.1 & 2068.93 MiB & 31.02\\
\bottomrule
\end{tabular}
\end{table}
\section{Conclusion}
This work presents, to our knowledge, the first application of an MLP-Mixer-like architecture to CPATH. Using a combination of feature embedding and clustering for preprocessing enables the use of this model with whole slide data. Although reducing the dataset and discarding much information, Mixer is still able to build good slide-level feature representations for cancer subtyping. 
Our implementation can compete with current state-of-the-art approaches such as CLAM on major benchmark datasets with competitive predictive performance, 24\% memory load and 15\% computational time reduction. 
One drawback of using our proposed method is the lack of interpretability: without attention scores or similar insight available, the decision reached by the model becomes more obtuse.

\subsubsection*{Acknowledgements}
This work was supported by JST CREST JPMJCR21D3 and Grant-in-Aid for Scientific Research (A) 23H00483. J.B. was supported by the Gateway Fellowship program of Research School, Ruhr-University Bochum, Bochum, Germany.
%Further research could be directed into improving clustering for a clearer separation of prototypes
%\section*{Acknowledgements}
\bibliographystyle{splncs04}
\bibliography{pathomix_bib}
\end{document}